\documentclass[10pt,twocolumn,letterpaper]{article}

\usepackage{cvpr}
\usepackage{times}
\usepackage{epsfig}
\usepackage{graphicx}
\usepackage{amsmath}
\usepackage{amssymb}
\usepackage{footnote}

\usepackage{algorithm}
\usepackage{algorithmic}
\usepackage{booktabs} 
\usepackage{bm}
\usepackage{mathtools}
\usepackage{makecell}
\usepackage{enumitem}
\usepackage{caption}
\usepackage{subcaption}
\usepackage{lipsum}
\usepackage{cuted}

\makeatletter
\DeclareRobustCommand\onedot{\futurelet\@let@token\@onedot}
\def\@onedot{\ifx\@let@token.\else.\null\fi\xspace}

\def\eg{\emph{e.g}\onedot} \def\Eg{\emph{E.g}\onedot}
\def\ie{\emph{i.e}\onedot} \def\Ie{\emph{I.e}\onedot}
 
\def\etc{\emph{etc}\onedot} \def\vs{\emph{vs}\onedot}
 
\def\etal{\emph{et al}\onedot}
\makeatother

\usepackage[breaklinks=true,bookmarks=false]{hyperref}

\cvprfinalcopy 

\def\cvprPaperID{3323} 

\ifcvprfinal\fi
\begin{document}

\title{Centripetal SGD for Pruning Very Deep Convolutional Networks with Complicated Structure \thanks{This work is supported by the National Key R\&D Program of China (No. 2018YFC0807500), National Natural Science Foundation of China (No. 61571269), National Postdoctoral Program for Innovative Talents (No. BX20180172), and the China Postdoctoral Science Foundation (No. 2018M640131). Corresponding author: Guiguang Ding.}}

\author{Xiaohan Ding \textsuperscript{1} \quad Guiguang Ding \textsuperscript{1} \quad Yuchen Guo \textsuperscript{1} \quad Jungong Han \textsuperscript{2}\\
	\textsuperscript{1} Tsinghua University	\quad	\textsuperscript{2}	Lancaster University\\
	\tt\small dxh17@mails.tsinghua.edu.cn dinggg@tsinghua.edu.cn \{yuchen.w.guo,jungonghan77\}@gmail.com
}

\maketitle
\thispagestyle{empty}

\begin{abstract}
The redundancy is widely recognized in Convolutional Neural Networks (CNNs), which enables to remove unimportant filters from convolutional layers so as to slim the network with acceptable performance drop. Inspired by the linear and combinational properties of convolution, we seek to make some filters increasingly close and eventually identical for network slimming. To this end, we propose Centripetal \footnote{Here ``centripetal'' means ``several objects moving towards a center'', not ``an object rotating around a center by the centripetal force''.} SGD (C-SGD), a novel optimization method, which can train several filters to collapse into a single point in the parameter hyperspace. When the training is completed, the removal of the identical filters can trim the network with \textit{NO} performance loss, thus no finetuning is needed. By doing so, we have partly solved an open problem of constrained filter pruning on CNNs with complicated structure, where some layers must be pruned following others. Our experimental results on CIFAR-10 and ImageNet have justified the effectiveness of C-SGD-based filter pruning. Moreover, we have provided empirical evidences for the assumption that the redundancy in deep neural networks helps the convergence of training by showing that a redundant CNN trained using C-SGD outperforms a normally trained counterpart with the equivalent width. 
\end{abstract}

\section{Introduction}
Convolutional Neural Network (CNN) has become an important tool for machine learning and many related fields \cite{collobert2008unified,lawrence1997face,lecun1995convolutional,lecun1990handwritten}. However, due to their nature of computational intensity, as CNNs grow wider and deeper, their memory footprint, power consumption and required floating-point operations (FLOPs) have increased dramatically, thus making them difficult to be deployed on platforms without rich computational resource, like embedded systems. In this context, CNN compression and acceleration methods have been intensively studied, including tensor low rank expansion \cite{jaderberg2014speeding}, connection pruning \cite{han2015learning}, filter pruning \cite{li2016pruning}, quantization \cite{han2015deep}, knowledge distillation \cite{hinton2015distilling}, fast convolution \cite{mathieu2013fast}, feature map compacting \cite{wang2017beyond}, \etc.

We focus on filter pruning, a.k.a. channel pruning \cite{he2017channel} or network slimming \cite{liu2017learning}, for three reasons. Firstly, filter pruning is a universal technique which is able to handle any kinds of CNNs, making no assumptions on the application field, the network architecture or the deployment platform. Secondly, filter pruning effectively reduces the FLOPs of the network, which serve as the main criterion of computational burdens. Lastly, as an important advantage in practice, filter pruning produces a thinner network with no customized structure or extra operation, which is orthogonal to the other model compression and acceleration techniques.

Motivated by the universality and significance, considerable efforts have been devoted to filter pruning techniques. Due to the widely observed redundancy in CNNs \cite{cheng2015exploration,collins2014memory,denil2013predicting,han2015deep,yu2018nisp,zhou2016less}, numerous excellent works have shown that, if a CNN is pruned appropriately with acceptable structural damage, a follow-up finetuning procedure can restore the performance to a certain degree. \textbf{1)} Some prior works \cite{abbasi2017structural,anwar2017structured,hu2016network,li2016pruning,molchanov2016pruning,polyak2015channel,yu2018nisp} sort the filters by their importance, directly remove the unimportant ones and re-construct the network with the remaining filters. As the important filters are preserved, a comparable level of performance can be reached by finetuning. However, some recent powerful networks have complicated structures, like identity mapping \cite{he2016deep} and dense connection \cite{huang2017densely}, where some layers must be pruned in the same pattern as others, raising an open problem of \textit{constrained filter pruning}. This further challenges such pruning techniques, as one cannot assume the important filters at different layers reside on the same positions. \textbf{2)} Obviously, the model is more likely to recover if the destructive impact of pruning is reduced. Taking this into consideration, another family of methods \cite{alvarez2016learning,ding2018auto,liu2015sparse,wang2018structured,wen2016learning} seeks to zero out some filters in advance, where group-Lasso Regularization \cite{roth2008group} is frequently used. Essentially, zeroing filters out can be regarded as producing a desired \textit{redundancy pattern} in CNNs. After reducing the magnitude of parameters of some whole filters, pruning these filters causes less accuracy drop, hence it becomes easier to restore the performance by finetuning.
\begin{figure*}
	\begin{center}
		\centerline{\includegraphics[width=\linewidth]{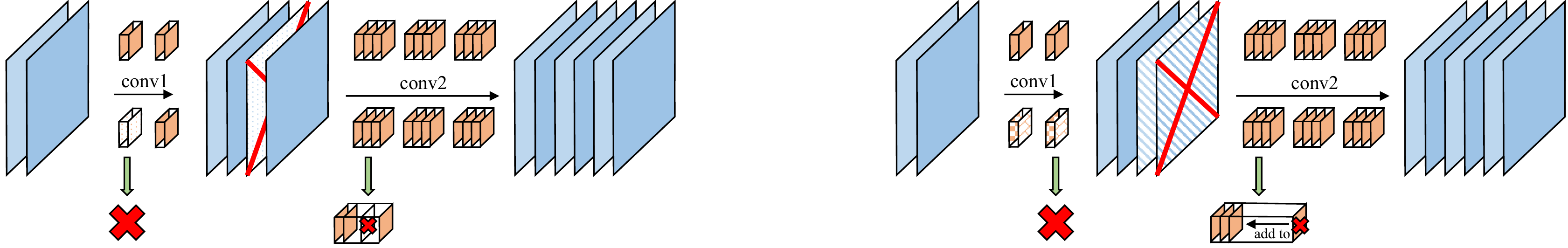}}
		\caption{Zeroing-out v.s. centripetal constraint. This figure shows a CNN with 4 and 6 filters at the 1st and 2nd convolutional layer, respectively, which takes a 2-channel input. Left: the 3rd filter at conv1 is zeroed out, thus the 3rd feature map is close to zero, implying that the 3rd input channels of the 6 filters at conv2 are useless. During pruning, the 3rd filters at conv1 along with the 3rd input channels of the 6 filters at conv2 are removed. Right: the 3rd and 4th filters at conv1 are forced to grow close by centripetal constraint until the 3rd and 4th feature maps become identical. But the 3rd and 4th input channels of the 6 filters at conv2 can still grow without constraints, making the encoded information still in full use. When pruned, the 4th filter at conv1 is removed, and the 4th input channel of every filter at conv2 is added to the 3rd channel.}
		\label{motivation-sketch}
	\end{center}
\vskip -0.25in
\end{figure*}

In this paper, we also aim to produce some redundancy patterns in CNNs for filter pruning. However, instead of zeroing out filters, which ends up with a pattern where some whole filters are close to zero, we intend to merge multiple filters into one, leading to a redundancy pattern where some filters are identical. The intuition motivating the proposed method is an observation of information flow in CNNs (Fig. \ref{motivation-sketch}). \textbf{1)} If two or more filters are trained to become identical, due to the \textit{linear} and \textit{combinational} properties of convolution, we can simply discard all but leave one filter, and add up the parameters along the corresponding input channels of the next layer. Doing so will cause ZERO performance loss, and there is no need for a time-consuming finetuning process. It is noted that such a finetuning process is essential for the zeroing-out methods \cite{alvarez2016learning,liu2015sparse,wen2016learning}, as the discarded filters are merely small in magnitude, but still encode a certain quantity of information. Therefore, removing such filters unavoidably degrades the performance of the network. \textbf{2)} When multiple filters are constrained to grow closer in the parameter hyperspace, which we refer to as the \textit{centripetal constraint}, though they start to produce increasingly similar information, the information conveyed from the corresponding input channels of the next layer is still in full use, thus the model's representational capacity is stronger than a counterpart with the filters being zeroed out.

We summarize our contributions as follows.
\begin{itemize}[noitemsep,nolistsep,,topsep=0pt,parsep=0pt,partopsep=0pt]
	\item We propose to produce redundancy patterns in CNNs by training some filters to become identical. Compared to the importance-based filter pruning methods, doing so requires no heuristic knowledge about the importance of filter. Compared to the zeroing-out methods, no finetuning is needed, and more representational capacity of the network is preserved. 
	\item We propose \textit{Centripetal SGD} (C-SGD), an innovative SGD optimization method. As the name suggests, we make multiple filters move towards a center in the hyperspace of the filter parameters. In the meantime, supervised by the model's original objective function, the performance is maintained as much as possible.
	\item By C-SGD, we have partly solved constrained filter pruning, an open problem of slimming modern very deep CNNs with complicated structure, where some layers must be pruned in the same pattern as others.
	\item We have presented both theoretical and empirical analysis of the effectiveness of C-SGD. We have shown empirical evidences supporting our motivation (Fig. \ref{motivation-sketch}) and the assumption that the redundancy helps the convergence of neural networks \cite{denton2014exploiting,hinton2015distilling}. The codes are available at \url{https://github.com/ShawnDing1994/Centripetal-SGD}.
\end{itemize}

\section{Related Work}
\textbf{Filter Pruning. }Numerous inspiring works \cite{castellano1997iterative,guo2016dynamic,han2015learning,hassibi1993second,lecun1990optimal,stepniewski1997pruning,zhang2018systematic} have shown that it is feasible to remove a large portion of connections or neurons from a neural network without a significant performance drop. However, as the connection pruning methods make the parameter tensors no smaller but just sparser, little or no acceleration can be observed without the support from specialized hardware. Then it is natural for researchers to go further on CNNs: by removing filters instead of sporadic connections, we transform the wide convolutional layers into narrower ones, hence the FLOPs, memory footprint and power consumption are significantly reduced. One kind of methods defines the importance of filters by some means, then selects and prunes the unimportant filters carefully to minimize the performance loss. Some prior works measure a filter's importance by the accuracy reduction (CAR) \cite{abbasi2017structural}, the channel contribution variance \cite{polyak2015channel}, the Taylor-expansion-based criterion \cite{molchanov2016pruning}, the magnitude of convolution kernels \cite{li2016pruning} and the average percentage of zero activations (APoZ) \cite{hu2016network}, respectively; Luo \etal \cite{luo2017thinet} select filters based on the information derived from the next layer; Yu \etal \cite{yu2018nisp} take into consideration the effect of error propagation; He \etal \cite{he2017channel} select filters by solving the Lasso regression; He and Han \cite{he2018adc} pick up filters with aid of reinforcement learning. Another category seeks to train the network under certain constraints in order to zero out some filters, where group-Lasso regularization is frequently used \cite{alvarez2016learning,liu2015sparse,wen2016learning}. It is noteworthy that since removing some whole filters can degrade the network a lot, the CNNs are usually pruned in a layer-by-layer \cite{alvarez2016learning,he2018adc,he2017channel,hu2016network,luo2017thinet,polyak2015channel} or filter-by-filter \cite{abbasi2017structural,molchanov2016pruning} manner, and require one or more finetuning processes to restore the accuracy \cite{abbasi2017structural,alvarez2016learning,anwar2017structured,he2018adc,he2017channel,hu2016network,li2016pruning,liu2017learning,luo2017thinet,molchanov2016pruning,polyak2015channel,wen2016learning,yu2018nisp}.

\textbf{Other Methods. }Apart from filter pruning, some excellent works seek to compress and accelerate CNNs in other ways. Considerable works \cite{alvarez2017compression,denton2014exploiting,jaderberg2014speeding,kim2015compression,sainath2013low,sindhwani2015structured,xue2013restructuring,zhang2016accelerating} decompose or approximate the parameter tensors; quantization and binarization techniques \cite{courbariaux2016binarized,gupta2015deep,han2015deep,rastegari2016xnor,wu2016quantized} approximate a model using fewer bits per parameter; knowledge distillation methods \cite{ba2014deep,hinton2015distilling,romero2014fitnets} transfer knowledge from a big network to a smaller one; some researchers seek to speed up convolution with the help of perforation \cite{figurnov2016perforatedcnns}, FFT \cite{mathieu2013fast,vasilache2014fast} or DCT \cite{wang2016cnnpack}; Wang \etal \cite{wang2017beyond} compact feature maps by extracting information via Circulant matrices. Of note is that since filter pruning simply shrinks a wide CNN into a narrower one with no special structures or extra operations, it is \textit{orthogonal} to the other methods.

\section{Slimming CNNs via Centripetal SGD} 
\subsection{Formulation}
In modern CNNs, batch normalization \cite{ioffe2015batch} and scaling transformation are commonly used to enhance the representational capacity of convolutional layers. For simplicity and generality, we regard the possible subsequent batch normalization and scaling layer as part of the convolutional layer. Let $i$ be the layer index, $\bm{M}^{(i)}\in\mathbb{R}^{h_{i}\times w_{i}\times c_{i}}$ be an $h_i\times w_i$ feature map with $c_{i}$ channels and $\bm{M}^{(i,j)}=\bm{M}^{(i)}_{:,:,j}$ be the $j$-th channel. The convolutional layer $i$ with kernel size $u_i\times v_i$ has one 4th-order tensor and four vectors as parameters at most, namely, $\bm{K}^{(i)}\in \mathbb{R}^{u_{i}\times v_{i}\times c_{i-1}\times c_{i}}$ and $\bm{\mu}^{(i)},\bm{\sigma}^{(i)},\bm{\gamma}^{(i)},\bm{\beta}^{(i)}\in\mathbb{R}^{c_{i}}$, where $\bm{K}^{(i)}$ is the convolution kernel, $\bm{\mu}^{(i)}$ and $\bm{\sigma}^{(i)}$ are the mean and standard deviation of batch normalization, $\bm{\gamma}^{(i)}$ and $\bm{\beta}^{(i)}$ are the parameters of the scaling transformation. Then we use $\bm{P}^{(i)}=(\bm{K}^{(i)},\bm{\mu}^{(i)},\bm{\sigma}^{(i)},\bm{\gamma}^{(i)},\bm{\beta}^{(i)})$ to denote the parameters of layer $i$.
In this paper, the filter $j$ at layer $i$ refers to the five-tuple comprising all the parameter slices related to the $j$-th output channel of layer $i$, formally, $\bm{F}^{(j)}=(\bm{K}^{(i)}_{:,:,:,j},\mu^{(i)}_j,\sigma^{(i)}_j,\gamma^{(i)}_j,\beta^{(i)}_j)$. During forward propagation, this layer takes $\bm{M}^{(i-1)}\in\mathbb{R}^{h_{i-1}\times w_{i-1}\times c_{i-1}}$ as input and outputs $\bm{M}^{(i)}$. Let $\ast$ be the 2-D convolution operator, the $j$-th output channel is given by
\begin{equation}\label{def-convolution}
\bm{M}^{(i,j)}=\frac{\sum_{k=1}^{c_{i-1}}\bm{M}^{(i-1,k)}\ast\bm{K}^{(i)}_{:,:,k,j}-\mu^{(i)}_j}{\sigma^{(i)}_j}\gamma^{(i)}_j+\beta^{(i)}_j \,.
\end{equation}

The importance-based filter pruning methods \cite{abbasi2017structural,hu2016network,li2016pruning,molchanov2016pruning,polyak2015channel,yu2018nisp} define the importance of filters by some means, prune the unimportant part and reconstruct the network using the remaining parameters. Let $\mathcal{I}_i$ be the filter index set of layer $i$ (\eg, $\mathcal{I}_2=\{1,2,3,4\}$ if the second layer has four filters), $T$ be the filter importance evaluation function and $\theta_i$ be the threshold. The remaining set, \ie, the index set of the filters which survive the pruning, is $\mathcal{R}_i=\{j\in \mathcal{I}_i \ |\ T(\bm{F}^{(j)})>\theta_i\}$. Then we reconstruct the network by assembling the parameters sliced from the original tensor or vectors of layer $i$ into the new parameters. That is,
\begin{equation}\label{eq4}
\hat{\bm{P}}^{(i)}=(\bm{K}^{(i)}_{:,:,:,\mathcal{R}_i},\bm{\mu}^{(i)}_{\mathcal{R}_i},\bm{\sigma}^{(i)}_{\mathcal{R}_i},\bm{\gamma}^{(i)}_{\mathcal{R}_i},\bm{\beta}^{(i)}_{\mathcal{R}_i}) \,.
\end{equation}
The input channels of the next layer corresponding to the pruned filters should also be discarded,  
\begin{equation}\label{eq6}
\hat{\bm{P}}^{(i+1)}=(\bm{K}^{(i+1)}_{:,:,\mathcal{R}_i,:},\bm{\mu}^{(i+1)},\bm{\sigma}^{(i+1)},\bm{\gamma}^{(i+1)},\bm{\beta}^{(i+1)}) \,.
\end{equation}

\subsection{Update Rule}
For each convolutional layer, we first divide the filters into clusters. The number of clusters equals the desired number of filters, as we preserve only one filter for each cluster. We use $\mathcal{C}_i$ and $\mathcal{H}$ to denote the set of all filter clusters of layer $i$ and a single cluster in the form of a filter index set, respectively. We generate the clusters evenly or by k-means \cite{hartigan1979algorithm}, between which our experiments demonstrate only minor difference (Table. \ref{exp-table-cifar}).
\begin{itemize}[noitemsep,nolistsep,,topsep=0pt,parsep=0pt,partopsep=0pt]
	\item \textbf{K-means clustering}. We aim to generate clusters with low intra-cluster distance in the parameter hyperspace, such that collapsing them into a single point less impacts the model, which is natural. To this end, we simply flatten the filter's kernel and use it as the feature vector for k-means clustering.
	\item \textbf{Even clustering}. We can generate clusters with no consideration of the filters' inherent properties. Let $c_i$ and $r_i$ be the number of original filters and desired clusters, respectively, then each cluster will have $\lceil c_i / r_i\rceil$ filters at most. For example, if the second layer has six filters and we wish to slim it to four filters, we will have $\mathcal{C}_2=\{\mathcal{H}_1,\mathcal{H}_2,\mathcal{H}_3,\mathcal{H}_4\}$, where $\mathcal{H}_1=\{1,2\},\mathcal{H}_2=\{3,4\},\mathcal{H}_3=\{5\},\mathcal{H}_4=\{6\}$.
\end{itemize}

We use $H(j)$ to denote the cluster containing filter $j$, so in the above example we have $H(3)=\mathcal{H}_2$ and $H(6)=\mathcal{H}_4$. Let $\bm{F}^{(j)}$ be the kernel or a vector parameter of filter $j$, at each training iteration, the update rule of C-SGD is
\begin{equation}\label{update-rule}
\begin{aligned}
\bm{F}^{(j)} \gets &\bm{F}^{(j)}+\tau \Delta\bm{F}^{(j)} \,, \\
\Delta\bm{F}^{(j)}=&-\frac{\sum_{k \in H(j)} \frac{\partial L}{\partial \bm{F}^{(k)}}}{|H(j)|} - \eta \bm{F}^{(j)} \\ &+ \epsilon (\frac{\sum_{k \in H(j)}\bm{F}^{(k)}}{|H(j)|}-\bm{F}^{(j)}) \,,
\end{aligned}
\end{equation}
where $L$ is the original objective function, $\tau$ is the learning rate, $\eta$ is the model's original weight decay factor, and $\epsilon$ is the only introduced hyper-parameter, which is called the \textit{centripetal strength}.

Let $\mathcal{L}$ be the layer index set, we use the \textit{sum of squared kernel deviation} $\chi$ to measure the intra-cluster similarity, \ie, how close filters are in each cluster,
\begin{equation}
\chi=\sum_{i\in\mathcal{L}}\sum_{j\in \mathcal{I}_i}||\bm{K}^{(i)}_{:,:,:,j} - \frac{\sum_{k\in H(j)}\bm{K}^{(i)}_{:,:,:,k}}{|H(j)|}||_2^2 \,.
\end{equation}
It is easy to derive from Eq. \ref{update-rule} that if the floating-point operation errors are ignored, $\chi$ is lowered \textit{monotonically} and \textit{exponentially} with a proper learning rate $\tau$.

The intuition behind Eq. \ref{update-rule} is quite simple: for the filters in the same cluster, the increments derived by the objective function are averaged (the first term), the normal weight decay is applied as well (the second term), and the difference in the initial values is gradually eliminated (the last term), so the filters will move towards their center in the hyperspace. 

In practice, we fix $\eta$ and reduce $\tau$ with time just as we do in normal SGD training, and set $\epsilon$ casually. Intuitively, C-SGD training with a large $\epsilon$ prefers ``rapid change'' to ``stable transition'', and vice versa. If $\epsilon$ is too large, \eg, 10, the filters are merged in an instant such that the whole process becomes equivalent to training a destroyed model from scratch. If $\epsilon$ is extremely small, like $1\times 10^{-10}$, the difference between C-SGD training and normal SGD is almost invisible during a long time. However, since the difference among filters in each cluster is reduced \textit{monotonically} and \textit{exponentially}, even an extremely small $\epsilon$ can make the filters close enough, sooner or later. As shown in the Appendix, C-SGD is insensitive to $\epsilon$.

\begin{figure}[t]\label{fig-contour}
	\begin{subfigure}{0.49\columnwidth}
		\includegraphics[width=\linewidth,height=1.50in]{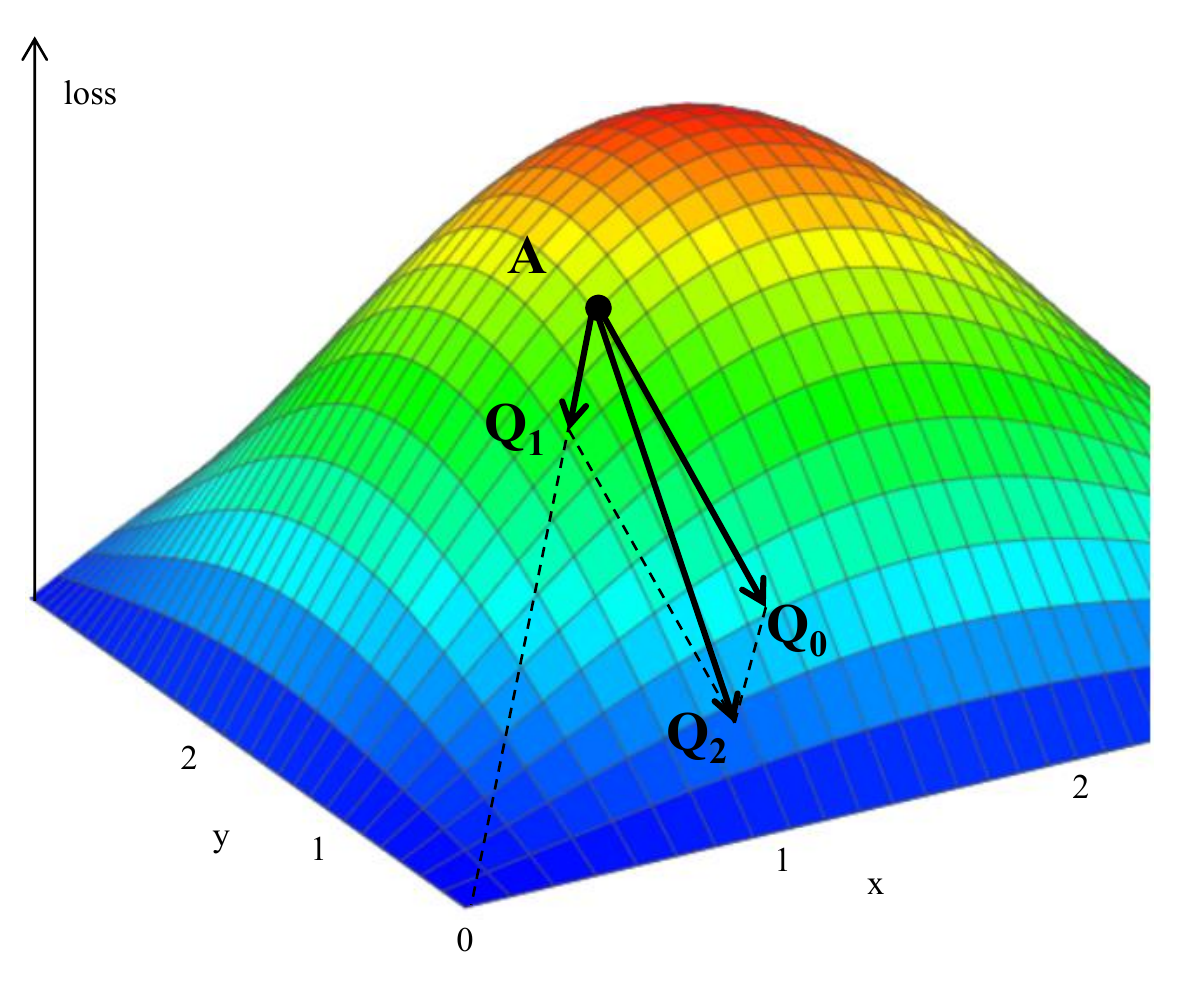} 
		\caption{Normal weight decay.}
		\label{fig-contour-weightdecay}
	\end{subfigure}
	\begin{subfigure}{0.49\columnwidth}
		\includegraphics[width=\linewidth,height=1.50in]{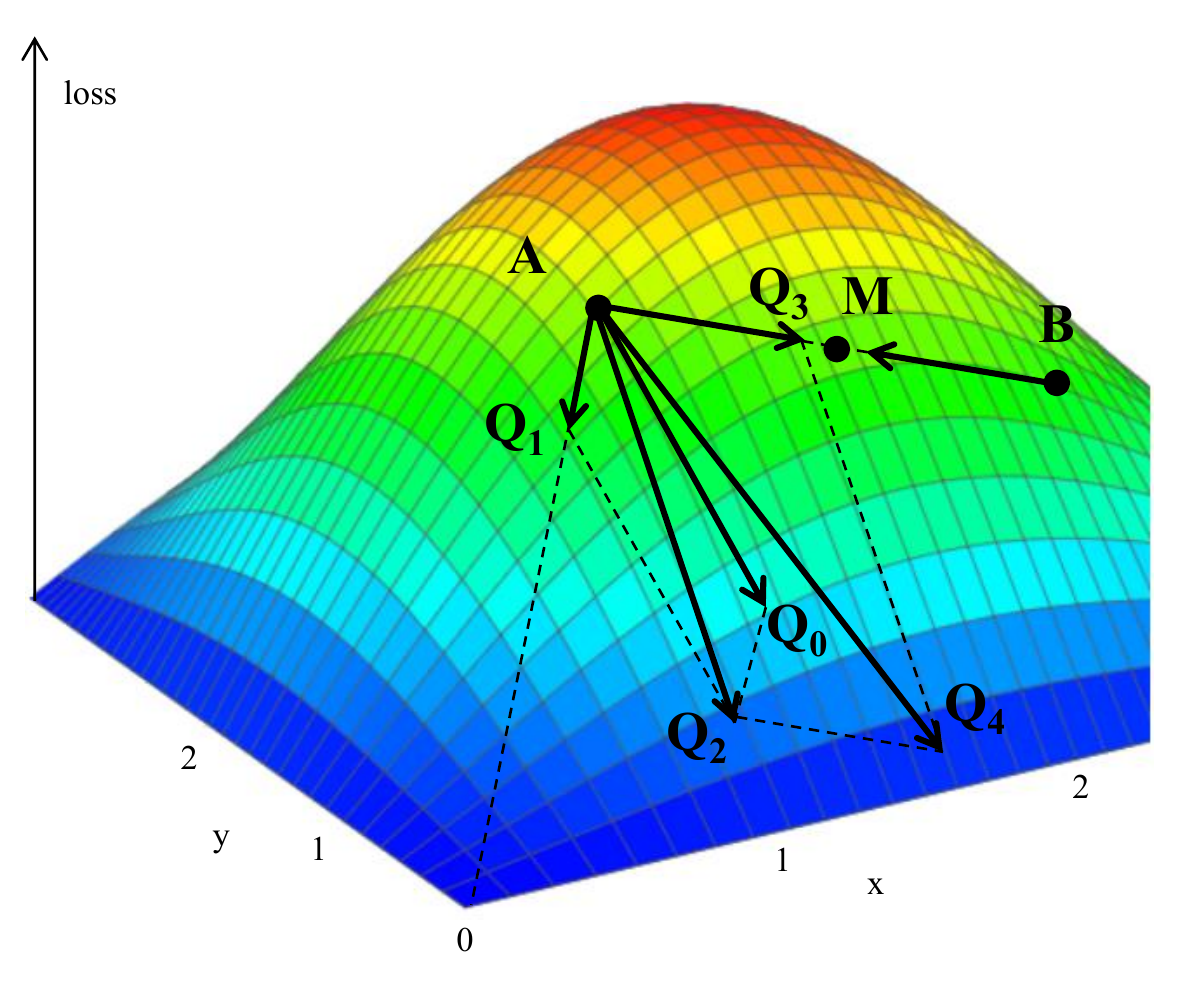}
		\caption{Centripetal constraint.}
		\label{fit-contour-csgd}
	\end{subfigure}
	\caption{Gradient descent direction on the loss surface of normal weight decay and centripetal constraint without merging the original gradients.}
\end{figure}
A simple analogy to weight decay (\ie, $\ell$-2 regularization) may help understand Centripetal SGD. Fig. \ref{fig-contour-weightdecay} shows a 3-D loss surface, where a certain point $A$ corresponds to a 2-D parameter $\bm{a}=(a_1,a_2)$. Suppose the steepest descent direction is $\overrightarrow{AQ_0}$, we have $\overrightarrow{AQ_0}=-\frac{\partial L}{\partial\bm{a}}$, where $L$ is the objective function. Weight decay is commonly applied to reduce overfitting \cite{krogh1992simple}, that is, $\overrightarrow{AQ_1}=-\eta\bm{a}$, where $\eta$ is the model's weight decay factor, \eg, $1\times 10^{-4}$ for ResNets \cite{he2016deep}. The actual gradient descent direction then becomes $\Delta\bm{a}=\overrightarrow{AQ_2}=\overrightarrow{AQ_0}+\overrightarrow{AQ_1}=-\frac{\partial L}{\partial\bm{a}}-\eta\bm{a}$.

Formally, with $t$ denoting the number of training iterations, we seek to make point $A$ and $B$ grow increasingly close and eventually the same by satisfying
\begin{equation}\label{lim-satisfy}
\lim\limits_{t\to\infty}||\bm{a}^{(t)}-\bm{b}^{(t)}||=0 \,.
\end{equation}
Given the fact that $\bm{a}^{(t+1)}=\bm{a}^{(t)}+\tau\Delta\bm{a}^{(t)}$ and $\bm{b}^{(t+1)}=\bm{b}^{(t)}+\tau\Delta\bm{b}^{(t)}$, where $\tau$ is the learning rate, Eq. \ref{lim-satisfy} implies
\begin{equation}
\lim\limits_{t\to\infty}||(\bm{a}^{(t)}-\bm{b}^{(t)})+\tau(\Delta\bm{a}^{(t)}-\Delta\bm{b}^{(t)})||=0 \,.
\end{equation}
We seek to achieve this with $\lim\limits_{t\to\infty}(\Delta\bm{a}^{(t)}-\Delta\bm{b}^{(t)})=\bm{0}$ as well as $\lim\limits_{t\to\infty}(\bm{a}^{(t)}-\bm{b}^{(t)})=\bm{0}$. Namely, as two points are growing closer, their gradients should become closer accordingly in order for the training to converge.

If we just wish to make $A$ and $B$ closer to each other than they used to be, a natural idea is to push both $A$ and $B$ to their midpoint $M(\frac{\bm{a} + \bm{b}}{2})$, as shown in Fig. \ref{fit-contour-csgd}. Therefore, the gradient descent direction of point $A$ becomes 
\begin{equation}\label{delta-a}
\Delta\bm{a}=\overrightarrow{AQ_2}+\overrightarrow{AQ_3}=-\frac{\partial L}{\partial\bm{a}}-\eta\bm{a}+\epsilon(\frac{\bm{a}+\bm{b}}{2}-\bm{a}) \,,
\end{equation}
where $\epsilon$ is a hyper-parameter controlling the intensity or speed of pushing $A$ and $B$ close. We have
\begin{equation}\label{delta-b}
\Delta\bm{b}=-\frac{\partial L}{\partial\bm{b}}-\eta\bm{b}+\epsilon(\frac{\bm{a}+\bm{b}}{2}-\bm{b}) \,,
\end{equation}
\begin{equation}
\Delta\bm{a}-\Delta\bm{b} = (\frac{\partial L}{\partial\bm{b}} -\frac{\partial L}{\partial\bm{a}}) + (\eta+\epsilon)(\bm{b}-\bm{a}) \,.
\end{equation}

Here we see the problem: we cannot ensure $\lim\limits_{t\to\infty}(\frac{\partial L}{\partial\bm{b}^{(t)}}-\frac{\partial L}{\partial\bm{a}^{(t)}})=\bm{0}$. Actually, even $\bm{a}=\bm{b}$ does not imply $\frac{\partial L}{\partial\bm{a}}=\frac{\partial L}{\partial\bm{b}}$, because they participate in different computation flows. As a consequence, we cannot ensure $\lim\limits_{t\to\infty}(\Delta\bm{a}^{(t)}-\Delta\bm{b}^{(t)})=\bm{0}$ with Eq. \ref{delta-a} and Eq. \ref{delta-b}. 

We solve this problem by merging the gradients derived from the original objective function. For simplicity and symmetry, by replacing both $\frac{\partial L}{\partial\bm{a}}$ in Eq. \ref{delta-a} and $\frac{\partial L}{\partial\bm{b}}$ in Eq. \ref{delta-b} with $\frac{1}{2}(\frac{\partial L}{\partial\bm{a}}+\frac{\partial L}{\partial\bm{b}})$, we have $\Delta\bm{a}-\Delta\bm{b} = (\eta+\epsilon)(\bm{b}-\bm{a})$. In this way, the supervision information encoded in the objective-function-related gradients is preserved to maintain the model's performance, and Eq. \ref{lim-satisfy} is satisfied, which can be easily verified. Intuitively, we deviate $\bm{a}$ from the steepest descent direction according to some information of $\bm{b}$ and deviate $\bm{b}$ vice versa, just like the $\ell$-2 regularization deviates both $\bm{a}$ and $\bm{b}$ towards the origin of coordinates. 

\subsection{Efficient Implementation of C-SGD}
The efficiency of modern CNN training and deployment platforms, \eg, Tensorflow \cite{abadi2016tensorflow}, is based on large-scale tensor operations. We therefore seek to implement C-SGD by efficient matrix multiplication which introduces minimal computational burdens. Concretely, given a convolutional layer $i$, the kernel $\bm{K}\in\mathbb{R}^{u_i\times v_i\times c_{i-1} \times c_i}$ and the gradient $\frac{\partial L}{\partial \bm{K}}$, we reshape $\bm{K}$ to $\bm{W}\in\mathbb{R}^{u_i v_i c_{i-1} \times c_i}$ and $\frac{\partial L}{\partial \bm{K}}$ to $\frac{\partial L}{\partial \bm{W}}$ accordingly. We construct the averaging matrix $\bm{\Gamma}\in\mathbb{R}^{c_i \times c_i}$ and decaying matrix $\bm{\Lambda}\in\mathbb{R}^{c_i \times c_i}$ as Eq. \ref{def-avg-matrix} and Eq. \ref{def-decay-matrix} such that Eq. \ref{matrix-mul} is equivalent to Eq. \ref{update-rule}, which can be easily verified. Obviously, when the number of clusters equals that of the filters, Eq. \ref{matrix-mul} degrades into normal SGD with $\bm{\Gamma}=diag(1), \bm{\Lambda}=diag(\eta)$. The other trainable parameters (\ie, $\bm{\gamma}$ and $\bm{\beta}$) are reshaped into $\bm{W}\in\mathbb{R}^{1 \times c_i}$ and handled in the same way. In practice, we observe almost no difference in the speed between normal SGD and C-SGD using Tensorflow on Nvidia GeForce GTX 1080Ti GPUs with CUDA9.0 and cuDNN7.0.
\begin{equation}\label{matrix-mul}
\bm{W} \gets \bm{W}-\tau (\frac{\partial L}{\partial \bm{W}}\bm{\Gamma} + \bm{W}\bm{\Lambda}) \,.
\end{equation}
\begin{equation}\label{def-avg-matrix}
\bm{\Gamma}_{m,n}=
\begin{dcases}
1/|H(m)| & \text{if $H(m)=H(n)$} \,, \\
0 & \text{elsewise} \,.
\end{dcases}
\end{equation}
\begin{equation}\label{def-decay-matrix}
\bm{\Lambda}_{m,n}=
\begin{dcases}
\eta + (1 - 1 / |H(m)|)\epsilon & \text{if $H(m)=H(n)$} \,, \\
0 & \text{elsewise} \,.
\end{dcases}
\end{equation}

\subsection{Filter Trimming after C-SGD}
After C-SGD training, since the filters in each cluster have become identical, as will be shown in Sect. \ref{sec-vs-zero-out}, picking up which one makes no difference. We simply pick up the first filter (\ie, the filter with the smallest index) in each cluster to form the remaining set for each layer, which is
\[
\mathcal{R}_i=\{min(\mathcal{H}) \ |\ \forall \mathcal{H} \in \mathcal{C}_i\}.
\]

For the next layer, we add the to-be-deleted input channels to the corresponding remaining one,
\[
\bm{K}^{(i+1)}_{:,:,k,:}\gets\sum \bm{K}^{(i+1)}_{:,:,H(k),:} \quad \forall k \in \mathcal{R}_i \,,
\]
then we delete the redundant filters as well as the input channels of the next layer following Eq. \ref{eq4}, \ref{eq6}. Due to the linear and combinational properties of convolution (Eq. \ref{def-convolution}), no damage is caused, hence \textit{no finetuning} is needed.

\subsection{C-SGD for Constrained Filter Pruning}
Recently, accompanied by the advancement of CNN design philosophy, several efficient and compact CNN architectures \cite{he2016deep,huang2017densely} have emerged and become favored in the real-world applications. Altough some excellent works \cite{hu2016network,kim2015compression,molchanov2016pruning,yu2018nisp,zhou2016less} have shown that the classical plain CNNs, \eg, AlexNet \cite{krizhevsky2012imagenet} and VGG \cite{simonyan2014very}, are highly redundant and can be pruned significantly, the pruned versions are usually still inferior to the more up-to-date and complicated CNNs in terms of both accuracy and efficiency.

We consider filter pruning for very deep and complicated CNNs challenging for three reasons. \textbf{1)} Firstly, these networks are designed in consideration of computational efficiency, which makes them inherently compact and efficient. \textbf{2)} Secondly, these networks are significantly deeper than the classical ones, thus the layer-by-layer pruning techniques become inefficient, and the errors can increase dramatically when propagated through multiple layers, making the estimation of filter importance less accurate \cite{yu2018nisp}. \textbf{3)} Lastly and most importantly, some innovative structures are heavily used in these networks, \eg, cross-layer connections \cite{he2016deep} and dense connections \cite{huang2017densely}, raising an open problem of constrained filter pruning. 

\begin{figure}
	\begin{subfigure}{0.24\columnwidth}
		\includegraphics[page=1,width=\linewidth]{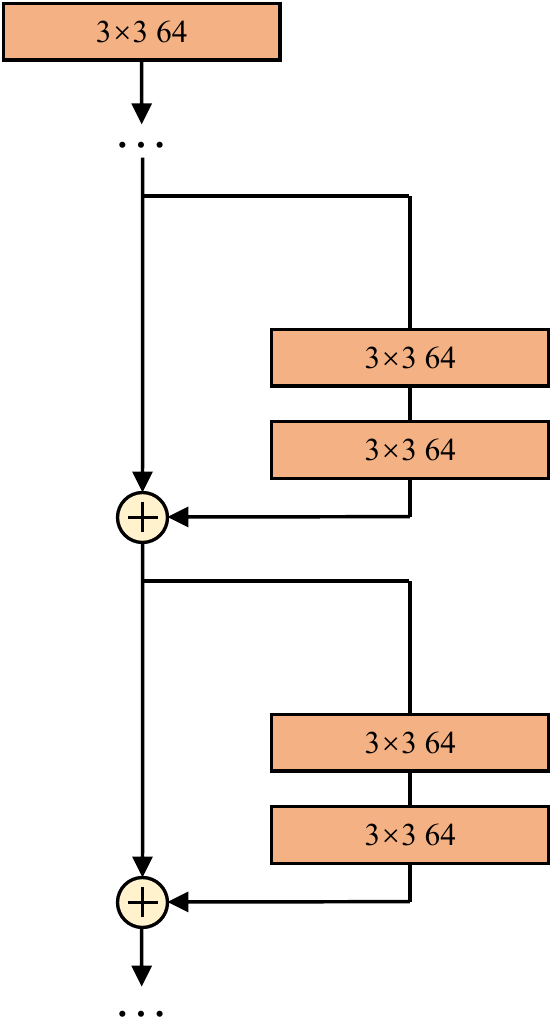} 
		\caption{Original.}
	\end{subfigure}
	\begin{subfigure}{0.24\columnwidth}
		\includegraphics[page=2,width=\linewidth]{literal_slim.pdf} 
		\caption{Clipped.}
	\end{subfigure}
	\begin{subfigure}{0.24\columnwidth}
		\includegraphics[page=3,width=\linewidth]{literal_slim.pdf} 
		\caption{Sampled.}
	\end{subfigure}
	\begin{subfigure}{0.24\columnwidth}
		\includegraphics[page=4,width=\linewidth]{literal_slim.pdf} 
		\caption{Slimmed.}
	\end{subfigure}
	\caption{Compared to the prior works which only clip the internal layers \cite{li2016pruning} or insert sampler layers \cite{he2017channel,liu2017learning} on ResNets, C-SGD is literally ``slimming'' the network.}
	\label{fig-slim-clip}
\end{figure}

\Ie, in each stage of ResNets, every residual block is expected to add the learned residuals to the stem feature maps produced by the first or the projection layer (referred to as \textit{pacesetter}), thus the last layer of every residual block (referred to as \textit{follower}) must be pruned in the same pattern as the pacesetter, \ie, the remaining set $\mathcal{R}$ of all the followers and the pacesetter must be identical, or the network will be damaged so badly that finetuning cannot restore its accuracy. For example, Li \etal. \cite{li2016pruning} once tried violently pruning ResNets but resulted in low accuracy. In some successful explorations, Li \etal \cite{li2016pruning} sidestep this problem by only pruning the internal layers on ResNet-56, \ie, the first layers in each residual block. Liu \etal \cite{liu2017learning} and He \etal \cite{he2017channel} skip pruning these troublesome layers and insert an extra sampler layer before the first layer in each residual block during inference time to reduce the input channels. Though these methods are able to prune the networks to some extent, from a holistic perspective the networks are not literally ``slimmed'' but actually ``clipped'', as shown in Fig. \ref{fig-slim-clip}.

We have partly solved this open problem by C-SGD, where the key is to force different layers to \textit{learn the same redundancy pattern}. For example, if the layer $p$ and $q$ have to be pruned in the same pattern, we only generate clusters for the layer $p$ by some means and assign the resulting cluster set to the layer $q$, namely, $\mathcal{C}_q\gets\mathcal{C}_p$. Then during C-SGD training, the same redundancy patterns among filters in both layer $p$ and $q$ are produced. \Ie, if the $j$-th and $k$-th filters at layer $p$ become identical, we ensure the sameness of the $j$-th and $k$-th filters at layer $q$ as well, thus the troublesome layers can be pruned along with the others. Some sketches are presented in the Appendix for more intuitions.

\section{Experiments}
\subsection{Slimming Very Deep and Complicated CNNs}\label{sec-exp1}
We experiment on CIFAR-10 \cite{krizhevsky2009learning} and ImageNet-1K \cite{deng2009imagenet} to evaluate our method. For each trial we start from a well-trained base model and apply C-SGD training on all the target layers \textit{simultaneously}. The comparisons between C-SGD and other filter pruning methods are presented in Table. \ref{exp-table-cifar} and Table. \ref{exp-table-standard} in terms of both absolute and relative error increase, which are commonly adopted as the metrics to fairly compare the change of accuracy on different base models. \Eg, the Top-1 accuracy of our ResNet-50 base model and C-SGD-70 is 75.33\% and 75.27\%, thus the absolute and relative error increase is $75.33\% - 75.27\% = 0.06\%$ and $\frac{0.06}{100-75.33}=0.24\%$, respectively.

\begin{table*}
	\setlength{\abovecaptionskip}{0pt} 
	\setlength{\belowcaptionskip}{0pt}
	\caption{Pruning Results on CIFAR-10. For C-SGD, the left is achieved by even clustering, and the right uses k-means.}
	\label{exp-table-cifar}
	\begin{center}
		\begin{small}
			\begin{tabular}{llcccccc}
				\toprule
				Model		&	Result 						&Base Top1	&\makecell{Pruned Top1 \\ even / k-means}						& \makecell{K-means Top1 error \\ Abs/Rel $\uparrow$\%}	& 	\makecell{FLOPs \\ $\downarrow$\%} 	&	Architecture		\\
				\midrule
				ResNet-56	&	Li \etal \cite{li2016pruning}			&	93.04	&	93.06			&	-0.02 / -0.28	&	27.60	&	only internals pruned \\
				ResNet-56	&	NISP-56 \cite{yu2018nisp}				&	-		&	-				&	0.03 / -		&	43.61	&	-	\\
				ResNet-56	&	Channel Pruning \cite{he2017channel}	&	92.8	&	91.8			&	1.0 / 13.88		&	50		&	sampler layer	\\
				ResNet-56	&	ADC \cite{he2018adc}					&	92.8	&	91.9			&	0.9	/ 12.5		&	50		&	sampler layer	\\
				\textbf{ResNet-56}	&	\textbf{C-SGD-5/8}&\textbf{93.39}	&\textbf{93.44 / 93.62}			&	\textbf{-0.23 / -3.47}	&	\textbf{60.85}	&	\textbf{10-20-40}	\\
				\midrule
				ResNet-110	&	Li \etal \cite{li2016pruning}			&	93.53	&	93.30			&	0.23 / 3.55		&	38.60	&	only internals pruned \\
				ResNet-110	&	NISP-110 \cite{yu2018nisp}				&	-		&	-				&	0.18 / -		&	43.78	&	-	\\
				\textbf{ResNet-110}	&	\textbf{C-SGD-5/8}&	\textbf{94.38}	&	\textbf{94.54 / 94.41}	&	\textbf{-0.03 / -0.53} 				&	\textbf{60.89}	&	\textbf{10-20-40}	\\
				\midrule
				ResNet-164	&	Network Slimming \cite{liu2017learning}&	94.58	&	94.73			&	-0.15 / -2.76	&	44.90	&	sampler layer	\\
				\textbf{ResNet-164}	&	\textbf{C-SGD-5/8}		&	\textbf{94.83}	&	\textbf{94.80 / 94.81}	&	\textbf{0.02 / 0.38}	&	\textbf{60.91}	&	\textbf{10-20-40}	\\	
				\midrule	
				DenseNet-40	&	Network Slimming \cite{liu2017learning}&	93.89	&	94.35			&	-0.46 / -7.52	&	55.00	&	-	\\
				\textbf{DenseNet-40}&	\textbf{C-SGD-5-8-10}	&	\textbf{93.81}	&	\textbf{94.37 / 94.56}			&	\textbf{-0.75 / 12.11}&	\textbf{60.05}	&	\textbf{5-8-10}	\\			
				\bottomrule
			\end{tabular}
		\end{small}
	\end{center}
\end{table*}

\begin{table*}
	\setlength{\abovecaptionskip}{0pt} 
	\setlength{\belowcaptionskip}{0pt}
	\caption{Pruning ResNet-50 on ImageNet using k-means clustering.}
	\label{exp-table-standard}
	\begin{center}
		\begin{small}
			\begin{tabular}{lccccccc}
				\toprule
				Result 							&Base Top1	&Base Top5	&Pruned Top1	&Pruned Top5	& \makecell{Top1 Error \\ Abs/Rel $\uparrow$\% }	&	\makecell{Top5 error \\ Abs/Rel $\uparrow$\% }	& 	\makecell{FLOPs \\ $\downarrow$\%} 			\\
				\midrule
				\textbf{C-SGD-70}	&	\textbf{75.33}	&	\textbf{92.56}	&	\textbf{75.27}		&	\textbf{92.46}		&	\textbf{0.06 / 0.24}  	&	\textbf{0.10 / 1.34}	&	\textbf{36.75}\\	
				ThiNet-70 \cite{luo2017thinet}	&72.88		&91.14		&72.04			&90.67			&0.84 / 3.09	&0.47 / 5.30	&36.75	\\
				SFP \cite{he2018soft}			&	76.15 	&	92.87	&	74.61		&	92.06		&	1.54 / 6.45		&	0.81 / 11.36	&	41.8\\
				NISP \cite{yu2018nisp}			&	-		&	-		&	-			&	-			&	0.89 / -		&	- / -			&	43.82\\	
				\textbf{C-SGD-60}		&	\textbf{75.33}	&	\textbf{92.56}	&	\textbf{74.93}		&	\textbf{92.27}		&	\textbf{0.40 / 1.62}	&	\textbf{0.29 / 3.89}	&	\textbf{46.24}	\\
				CFP \cite{singh2018leveraging}	&	75.3	&	92.2	&	73.4		&	91.4		&	1.9 / 7.69		&	0.8 / 10.25		&	49.6\\
				Channel Pruning \cite{he2017channel}	&	- 		&	92.2	&	-			&	90.8		&	- / -			&	1.4 / 17.94		&	50\\
				Autopruner \cite{luo2018autopruner}	&	76.15 	&	92.87	&	74.76		&	92.15		&	1.39 / 5.82		&	0.72 / 10.09	&	51.21\\
				GDP \cite{lin2018accelerating}	&	75.13 	&	92.30	&	71.89		&	90.71		&	3.24 / 13.02	&	1.59 / 20.64	&	51.30\\
				SSR-L2 \cite{lin2019towards}	&	75.12 	&	92.30	&	71.47		&	90.19		&	3.65 / 14.67	&	2.11 / 27.40	&	55.76\\	
				DCP \cite{zhuang2018discrimination}&76.01 	&	92.93	&	74.95		&	92.32		&	1.06 / 4.41		&	0.61 / 8.62		&	55.76\\	
				ThiNet-50 \cite{luo2017thinet}						&72.88		&91.14		&71.01			&90.02			&1.87 / 6.89	&1.12 / 12.64	&55.76	\\	
				\textbf{C-SGD-50}				&	\textbf{75.33}	&	\textbf{92.56}	&	\textbf{74.54}		&	\textbf{92.09}		&	\textbf{0.79 / 3.20}	&	\textbf{0.47 / 6.31}	&	\textbf{55.76}\\		
				\bottomrule
			\end{tabular}
		\end{small}
	\end{center}
\end{table*}

\textbf{CIFAR-10.} The base models are trained from scratch for 600 epochs to ensure the convergence, which is much longer than the usually adopted benchmarks (160 \cite{he2016deep} or 300 \cite{huang2017densely} epochs), such that the improved accuracy of the pruned model cannot be simply attributed to the extra training epochs on a base model which has not fully converged. We use the data augmentation techniques adopted by \cite{he2016deep}, \ie, padding to $40\times40$, random cropping and flipping. The hyper-parameter $\epsilon$ is casually set to $3\times10^{-3}$. We perform C-SGD training with batch size 64 and a learning rate initialized to $3\times10^{-2}$ then decayed by 0.1 when the loss stops decreasing. For each network we perform two experiments independently, where the only difference is the way we generate filter clusters, namely, even dividing or k-means clustering. We seek to reduce the FLOPs of every model by around 60\%, so we prune $3/8$ of \textit{every} convolutional layer of ResNets, thus the parameters and FLOPs are reduced by around $1-(5/8)^2=61\%$. Aggressive as it is, no obvious accuracy drop is observed. For DenseNet-40, the pruned model has 5, 8 and 10 incremental convolutional layers in the three stages, respectively, so that the FLOPs is reduced by 60.05\%, and a significantly increased accuracy is observed, which is consistent with but better than that of \cite{liu2017learning}.

\textbf{ImageNet.} We perform experiments using ResNet-50 \cite{he2016deep} on ImageNet to validate the effectiveness of C-SGD on the real-world applications. We apply k-means clustering on the filter kernels to generate the clusters, then use the ILSVRC2015 training set which contains 1.28M high-quality images for training. We adopt the standard data augmentation techniques including b-box distortion and color shift. At test time, we use a single central crop. For C-SGD-7/10, C-SGD-6/10 and C-SGD-5/10, all the first and second layers in each residual block are shrunk to 70\%, 60\% and 50\% of the original width, respectively.

\textbf{Discussions.} Our pruned networks exhibit fewer FLOPs, simpler structures and higher or comparable accuracy. Note that we apply the same pruning ratio globally for ResNets, and better results are promising to be achieved if more layer sensitivity analyzing experiments \cite{he2017channel,li2016pruning,yu2018nisp} are conducted, and the resulting network structures are tuned accordingly.  Interestingly, even arbitrarily generated clusters can produce reasonable results (Table. \ref{exp-table-cifar}).

\subsection{Redundant Training \vs Normal Training}\label{sec-exp2}
The comparisons between C-SGD and other pruning-and-finetuning methods \cite{he2017channel,li2016pruning,luo2017thinet,yu2018nisp} indicate that it may be better to train a redundant network and equivalently transform it to a narrower one than to finetune it after pruning. This observation is consistent with \cite{denton2014exploiting} and \cite{hinton2015distilling}, where the authors believe that the redundancy in neural networks is necessary to overcome a highly non-convex optimization. 

We verify this assumption by training a narrow CNN with normal SGD and comparing it with another model trained using C-SGD with the \textit{equivalent width}, which means that some redundant filters are produced during training and trimmed afterwards, resulting in the same network structure as the normally trained model. For example, if a network has $2\times$ number of filters as the normal counterpart but every two filters are identical, they will end up with the same structure. If the redundant one outperforms the normal one, we can conclude that C-SGD does yield more powerful networks by exploiting the redundant filters.

On DenseNet-40, we evenly divide the 12 filters at each incremental layer into 3 clusters, use C-SGD to train the network from scratch, then trim it to obtain a DenseNet-40 with 3 filters per incremental layer. \Ie, during training, every 4 filters are growing centripetally. As contrast, we train a DenseNet-40 with originally 3 filters per layer by normal SGD. Another group of experiments where each layer ends up with 6 filters are carried out similarly. After that, experiments on VGG \cite{simonyan2014very} are also carried out, where we slim each layer to 1/4 and 1/2 of the original width, respectively. It can be concluded from Table. \ref{exp-table-redundant-normal} that the redundant filters do help, compared to a normally trained counterpart with the equivalent width. This observation supports our intuition that the centripetally growing filters can maintain the model's representational capacity to some extent because though these filters are constrained, their corresponding input channels are still in full use and can grow without constraints (Fig. \ref{motivation-sketch}).
\begin{table}[t]
	\setlength{\abovecaptionskip}{0pt} 
	\setlength{\belowcaptionskip}{0pt}
	\caption{Validation accuracy of scratch-trained DenseNet-40 and VGG using C-SGD or normal SGD on CIFAR-10.}
	\label{exp-table-redundant-normal}
	\begin{center}
		\begin{small}
			\begin{tabular}{lcc}
				\toprule
				Model	   	& Normal SGD 	&C-SGD	\\
				\midrule
				DenseNet-3	&88.60	&\textbf{89.96}	\\
				DenseNet-6	&89.96	&\textbf{90.89}	\\
				VGG-1/4		&90.16	&\textbf{90.64}	\\
				VGG-1/2		&92.49	&\textbf{93.22}	\\
				\bottomrule
			\end{tabular}
		\end{small}
	\end{center}
\end{table}

\subsection{Making Filters Identical \vs Zeroing Out}\label{sec-vs-zero-out}
As making filters identical and zeroing filters out \cite{alvarez2016learning,ding2018auto,liu2015sparse,wang2018structured,wen2016learning} are two means of producing redundancy patterns for filter pruning, we perform controlled experiments on ResNet-56 to investigate the difference. For fair comparison, we aim to produce the same number of redundant filters in both the model trained with C-SGD and the one with group-Lasso Regularization \cite{roth2008group}. For C-SGD, the number of clusters in each layer is 5/8 of the number of filters. For Lasso, 3/8 of the original filters in the pacesetters and internal layers are regularized by group-Lasso, and the followers are handled in the same pattern. We use the aforementioned sum of squared kernel deviation $\chi$ and the \textit{sum of squared kernel residuals} $\phi$ as follows to measure the redundancy, respectively. Let $\mathcal{L}$ be the layer index set and $\mathcal{P}_i$ be the to-be-pruned filter set of layer $i$, \ie, the set of the 3/8 filters with group-Lasso regularization,
\[
\quad\quad \phi=\sum_{i\in\mathcal{L}}\sum_{j\in \mathcal{P}_i}||\bm{K}^{(i)}_{:,:,:,j}||_2^2 \,.
\]

\begin{figure}[t]
	\begin{subfigure}{0.49\columnwidth}
		\includegraphics[width=\linewidth]{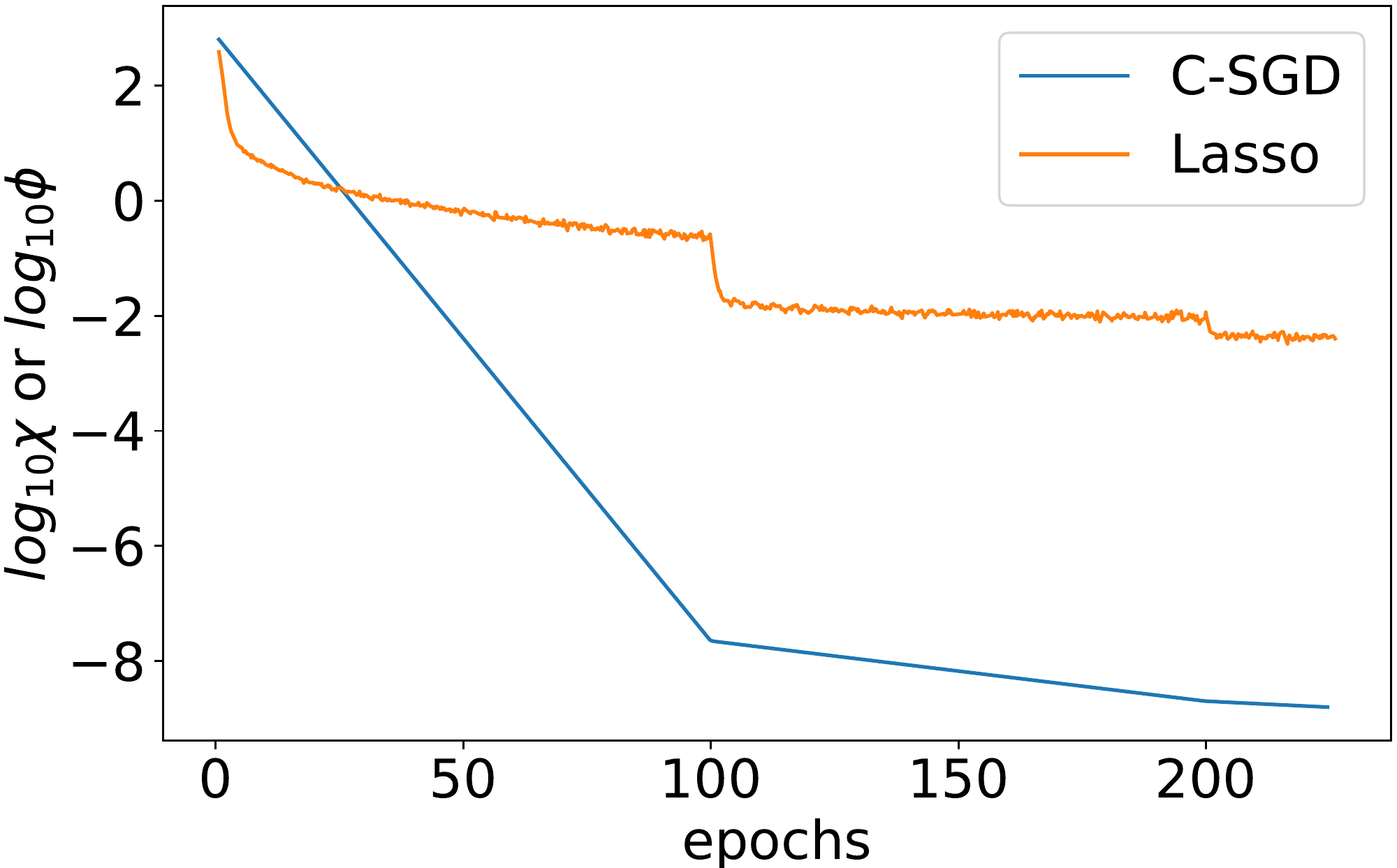} 
		\caption{Values of $\chi$ or $\phi$.}
		\label{curve-chi-phi}
	\end{subfigure}
	\begin{subfigure}{0.49\columnwidth}
		\includegraphics[width=\linewidth]{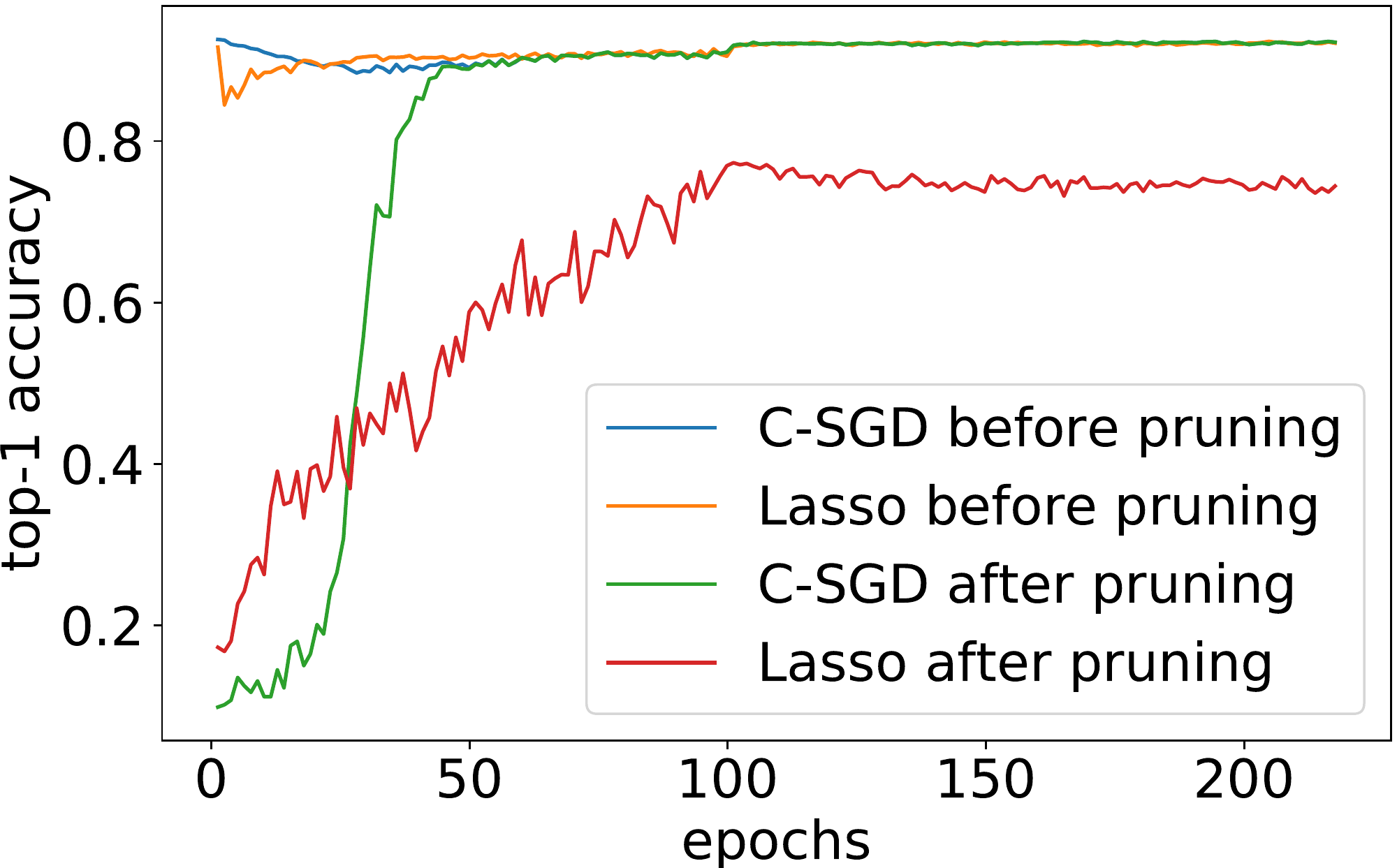}
		\caption{Validation accuracy.}
		\label{curve-acc-epochs}
	\end{subfigure}
	\caption{Training process with C-SGD or group-Lasso on ResNet-56. Note the logarithmic scale of the left figure.}
	\label{fig-chi-phi-acc-before-and-after}
\end{figure}

We present in Fig. \ref{fig-chi-phi-acc-before-and-after} the curves of $\chi$, $\phi$ as well as the validation accuracy both before and after pruning. The learning rate $\tau$ is initially set to $3\times10^{-2}$ and decayed by 0.1 at epoch 100 and 200, respectively. It can be observed that: \textbf{1)} Group Lasso cannot literally zero out filters, but can decrease their magnitude to some extent, as $\phi$ plateaus when the gradients derived from the regularization term become close to those derived from the original objective function. We empirically find out that even when $\phi$ reaches around $4\times10^{-4}$, which is nearly $2\times10^{6}$ times smaller than the initial value, pruning still causes obvious damage (around 10\% accuracy drop). When the learning rate is decayed and $\phi$ is reduced at epoch 200, we observe no improvement in the pruned accuracy, therefore no more experiments with smaller learning rate or stronger group-Lasso regularization are conducted. We reckon this is due to the error propagation and amplification in very deep CNNs \cite{yu2018nisp}. \textbf{2)} By C-SGD, $\chi$ is reduced \textit{monotonically} and perfectly \textit{exponentially}, which leads to faster convergence. \Ie, the filters in each cluster can become \textit{infinitely close} to each other at a \textit{constant rate} with a constant learning rate. For C-SGD, pruning causes \textit{absolutely no} performance loss after around 90 epochs. \textbf{3)} Training with group-Lasso is $2\times$ slower than C-SGD as it requires costly square root operations.

\subsection{C-SGD \vs Other Filter Pruning Methods}\label{sec-exp3}
\begin{figure}[t]\label{fig-compare-prune}
	\begin{subfigure}{0.49\columnwidth}
		\includegraphics[width=\linewidth,height=1.50in]{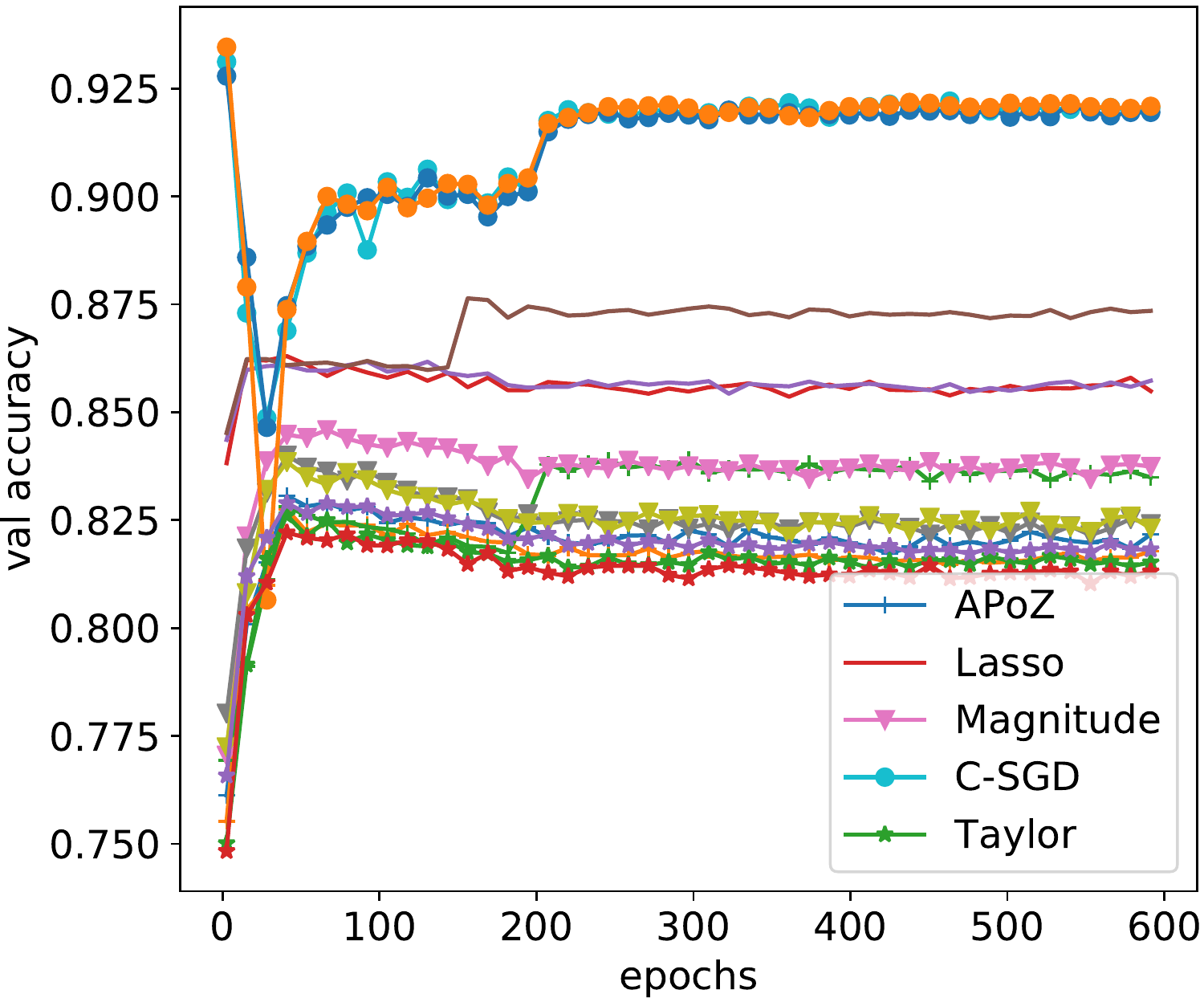} 
		\caption{Three filters per layer.}
		\label{densenet40-3}
	\end{subfigure}
	\begin{subfigure}{0.49\columnwidth}
		\includegraphics[width=\linewidth,height=1.50in]{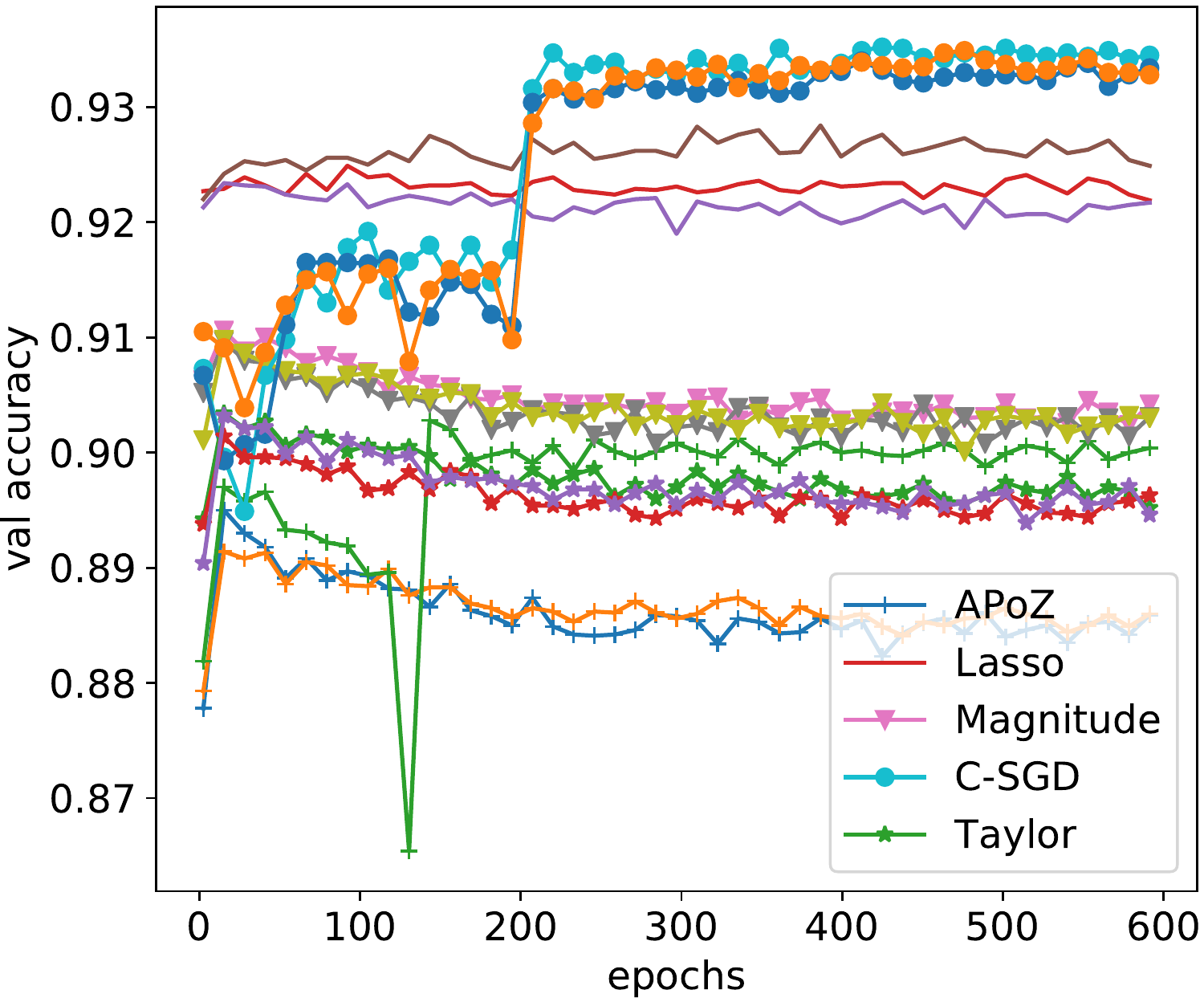}
		\caption{Six filters per layer.}
		\label{densenet40-6}
	\end{subfigure}
	\caption{Controlled pruning experiments on DenseNet-40.}
	\label{pruning-dc40}
\end{figure}
We compare C-SGD with other methods by controlled experiments on DenseNet-40 \cite{huang2017densely}. We slim \textit{every} incremental layer of a well-trained DenseNet-40 to 3 and 6 filters, respectively. The experiments are repeated 3 times, and all the results are presented in Fig. \ref{pruning-dc40}. The training setting is kept the same for every model: learning rate $\tau=3\times 10^{-3},3\times 10^{-4},3\times 10^{-5},3\times 10^{-6}$ for 200, 200, 100 and 100 epochs, respectively, to ensure the convergence of every model. For our method, the models are trained with C-SGD and trimmed. For Magnitude- \cite{li2016pruning}, APoZ- \cite{hu2016network} and Taylor-expansion-based \cite{molchanov2016pruning}, the models are pruned by different criteria and finetuned. The models labeled as Lasso are trained with group-Lasso Regularization for 600 epochs in advance, pruned, then finetuned for \textit{another} 600 epochs with the same learning rate schedule, so that the comparison is actually biased towards the Lasso method. The models are tested on the validation set every 10,000 iterations (12.8 epochs). The results reveal the superiority of C-SGD in terms of higher accuracy and also the better stability. Though group-Lasso Regularization can indeed reduce the performance drop caused by pruning, it is outperformed by C-SGD by a large margin. It is interesting that the violently pruned networks are unstable and easily trapped in the local minimum, \eg, the accuracy curves increase steeply in the beginning but slightly decline afterwards. This observation is consistent with that of Liu \etal \cite{liu2018rethinking}.

\section{Conclusion}
We have proposed to produce identical filters in CNNs for network slimming. The intuition is that making filters identical can not only eliminate the need for finetuning but also preserve more representational capacity of the network, compared to the zeroing-out fashion (Fig. \ref{motivation-sketch}). We have partly solved an open problem of constrained filter pruning on very deep and complicated CNNs and achieved state-of-the-art results on several common benchmarks. By training networks with redundant filters using C-SGD, we have demonstrated empirical evidences for the assumption that redundancy can help the convergence of neural network training, which may encourage future studies. Apart from pruning, we consider C-SGD promising to be applied as a means of regularization or training technique.

{\small
\bibliographystyle{ieee}
\bibliography{crbib}
}
\end{document}